\documentclass[conference]{IEEEtran}
\IEEEoverridecommandlockouts
\usepackage{cite}
\usepackage{amsmath,amssymb,amsfonts}
\usepackage{algorithmic}
\usepackage{hyperref}
\usepackage{graphicx}
\usepackage[shortcuts,acronym]{glossaries}
\usepackage{textcomp}
\usepackage{enumitem}
\usepackage{booktabs}       
\usepackage[dvipsnames]{xcolor}
\usepackage{caption}
\usepackage{subcaption}
\usepackage{algorithm}
\usepackage{algorithmic}
\usepackage{svg}

\newacronym{dl}{DL}{Deep Learning}
\newacronym{rl}{RL}{Reinforcement Learning}
\newacronym{mrl}{Meta-RL}{Meta-Reinforcement Learning}
\newacronym{drl}{DRL}{Deep Reinforcement Learning}
\newacronym{ood}{OOD}{Out-of-Distribution}
\newacronym{fmcw}{FMCW}{Frequency Modulated Continuous Wave}
\newacronym{rai}{RAI}{Range-Angle Image}
\newacronym{mdp}{MDP}{Markov Decision Process}
\newacronym{ukf}{UKF}{Unscented Kalman Filter}
\newacronym{aoa}{AoA}{Angle of Arrival}
\newacronym{cfar}{CFAR}{Constant False Alarm Rate}
\newacronym{ml}{ML}{Machine Learning}
\newacronym{rmse}{RMSE}{Root Mean Squared Error}
\newacronym{fft}{FFT}{Fast-Fourier Transformation}
\def\BibTeX{{\rm B\kern-.05em{\sc i\kern-.025em b}\kern-.08em
    T\kern-.1667em\lower.7ex\hbox{E}\kern-.125emX}}
\begin{document}

\title{Uncertainty-based Meta-Reinforcement Learning for Robust Radar Tracking\\
}
\makeatletter
\newcommand{\linebreakand}{%
  \end{@IEEEauthorhalign}
  \hfill\mbox{}\par
  \mbox{}\hfill\begin{@IEEEauthorhalign}
}
\makeatother

\author{\IEEEauthorblockN{1\textsuperscript{st} Julius Ott}
\IEEEauthorblockA{\textit{Infineon Technologies AG}\\
julius.ott@infineon.com}
\and
\IEEEauthorblockN{2\textsuperscript{nd} Lorenzo Servadei}
\IEEEauthorblockA{\textit{Infineon Technologies AG}\\
lorenzo.servadei@infineon.com}
\and
\IEEEauthorblockN{3\textsuperscript{rd} Gianfranco Mauro}
\IEEEauthorblockA{\textit{Infineon Technologies AG}\\
gianfranco.mauro@infineon.com}
\linebreakand
\IEEEauthorblockN{4\textsuperscript{th} Thomas Stadelmayer}
\IEEEauthorblockA{\textit{Infineon Technologies AG}\\
thomas.stadelmayer@infineon.com}
\and
\IEEEauthorblockN{5\textsuperscript{th} Avik Santra}
\IEEEauthorblockA{\textit{Infineon Technologies AG}\\
avik.santra@infineon.com}
\and
\IEEEauthorblockN{6\textsuperscript{th} Robert Wille}
\IEEEauthorblockA{\textit{Technichal University of Munich}\\
robert.wille@tum.de}
}
\maketitle

\begin{abstract}
Nowadays, \ac{dl} methods often overcome the limitations of traditional signal processing approaches. Nevertheless, \ac{dl} methods are barely applied in real-life applications. This is mainly due to limited robustness and distributional shift between training and test data. To this end, recent work has proposed uncertainty mechanisms to increase their reliability. Besides, meta-learning aims at improving the generalization capability of \ac{dl} models. By taking advantage of that, this paper proposes an uncertainty-based \ac{mrl} approach with \ac{ood} detection. The presented method performs a given task in unseen environments and provides information about its complexity. This is done by determining first and second-order statistics on the estimated reward. Using information about its complexity, the proposed algorithm is able to point out when tracking is reliable. To evaluate the proposed method, we benchmark it on a radar-tracking dataset.
There, we show that our method outperforms related \ac{mrl} approaches on unseen tracking scenarios in peak performance by $\mathbf{16}$\% and the baseline by $\mathbf{35}$\% while detecting \ac{ood} data with an F1-Score of $\mathbf{72}$\%. This shows that our method is robust to environmental changes and reliably detects \ac{ood} scenarios.
\end{abstract}


\section{Introduction} \label{sec:intro}

Radar sensors are gaining momentum in the modern semiconductor industry. Various modulation types, independence of light conditions, low-cost and privacy-friendly features lead the radar technology to be successfully employable in applications such as people detection and object tracking \cite{ servadei2022label}. To keep up the pace of advancements, attention directs to applications like multi-person tracking, which is critical in several areas such as automotive safety, medical services, or logistics \cite{automotive, medical}.
Multi-person tracking attempts to estimate the position of each target in a scene. For estimating the track positions, the \ac{ukf} is often used in radar tasks \cite{kalman}. This method has the scope of a bayesian estimation of the target position incorporating a nonlinear dynamic movement model. The nonlinear transition is approximated by the unscented transform, described in \cite{unscentedtrans}. By design, the \ac{ukf} relies on hyperparameters. These tunable hyperparameters describe the dynamics of the sensor system and real-world scenarios. Particularly for radar data, occlusions, non-human disturbances, and limited resolution are significant challenges for robust tracking. Thus, the choice of optimal hyperparameters varies with the environment, and often their initial choice is suboptimal. Given the environment's variety of settings and noise, finding the best hyperparameters for any possible scenario is infeasible.  \newline
Recent work focused on estimating the underlying dynamics of a \ac{ukf} with specific neural networks \cite{kalmannet} or using \ac{rl} to provide the best hyperparameters for a given scene, as shown in \cite{ wang2020reinforcement, li2020adaptive}. Although this approach is promising, it often lacks robustness, and the data distribution on inference time might be different from the one used for training. Furthermore, the \ac{ukf} model might fail to track in overcrowded scenarios, given the capability of its underlying model. \newline 
In real-life applications such as automotive radar sensing, robustness also has to be critical, and then \ac{drl} is employed. However, neural networks are known to fail when the test data distribution is far off the training data distribution \cite{abiodun2018state}. As a consequence, meta-learning is often used in literature to close the gap between training and test data distributions \cite{andrychowicz2016learning} and aims to adapt quickly to novel tasks. 
Although model-based meta-learning algorithms have obtained excellent results, they are limited to the network design stage \cite{snail}.
For this reason, research has recently focused on model-agnostic meta-learning, which enables learning tasks independently of the machine learning model. This is granted through task-specific optimization and is widely applicable in domains such as few-shot classification and \ac{rl}. In the case of \ac{mrl}, context variables are a promising way to incorporate task-specific information, as shown in \cite{pearl}. In this method, the context variable is learned by a neural network from task-specific data. Although the method is performant, it requires computing a context variable using multiple data samples from each additional task. Storing this data during inference is inefficient. Hence, our proposed method uses domain information to formulate a context variable. Our approach does not need to store data: it computes the context variable from easy-to-obtain input distribution statistics, which we refer to as context prior. Furthermore, as shown in the experiments, our method leads to an improved domain generalization compared to other state-of-the-art approaches.
Nevertheless, in several real-world applications (e.g., radar-tracking), improved domain generalization cannot address the limitations inherent to the task itself. For example, tracking operations in small, crowded scenarios with obstacles becomes increasingly tricky. In order to assess the reliability of our \ac{mrl} method, we develop an uncertainty mechanism via bootstrapped networks. The uncertainty mechanism is combined with the context prior that encodes information about the task difficulty. In this approach, scenes where tracking is prone to failure, are classified as \ac{ood}, thus assessing the particular reliability of the tracker in the current scenario.\newline
\noindent As a summary, the contributions of the paper are the followings: 
\begin{enumerate}[nolistsep]
    \item \ac{mrl} for domain generalization without additional memory footprint using context priors
    \item Enhanced \ac{ood} detection with context priors that encode task difficulty
\end{enumerate} 
The remainder of the paper proceeds as follows: in Section \ref{sec:back}, we introduce the radar-tracking problem and how to tackle it with \ac{rl}. Afterward, we explain the specific signal processing in Section \ref{sec:approach}. In the same section, we show how the input data distribution is used to compute an informative context variable. At the end of this section, using the context variable, we propose a \ac{mrl} algorithm for environment generalization and detection of \ac{ood} scenarios. Finally, we evaluate our method on a multi-target radar tracking dataset against related \ac{mrl} methods in Section \ref{sec:eval}. Our proposed approach outperforms comparable \ac{mrl} approaches in terms of peak performance by $16$\% on the test scenarios and the baseline of fixed parameters by $35$\%. In the same way, it detects \ac{ood} scenarios with an F1-score of $72$\%. Thus, our approach is more robust to environmental changes and reliably detects \ac{ood} scenarios.\newline
In Section \ref{sec:results}, we summarize our results and give an outlook on future work in Section.

\section{Background and Motivation} \label{sec:back}
In this section, we review the background and related work. In Section \ref{subsec:tracking}, we first outline the principle of radar tracking. Afterward, we explain how \ac{rl} can be used to optimize radar tracking. Additionally, we extend this concept by introducing the fundamentals of \ac{mrl} and Uncertainty-based \ac{rl}. 
\subsection{Radar Tracking} \label{subsec:tracking}
\ac{fmcw} radars can estimate the range, \ac{aoa}, and velocity of targets. In the case of radar tracking, we use the range and \ac{aoa} to determine the target position.
The typical radar tracking pipeline can be divided into signal processing, detection, clustering, and tracking, as shown in \cite{santra2020deep}. A high level description is given in Figure \ref{fig:track_pipe}. The signal processing stage elaborates the sensor data from each radar antenna to estimate the reflected signal's range and angle. The resulting image is a so-called \ac{rai}. Afterward, the \ac{rai} is convolved with a window that determines the signal threshold based on the surrounding noise. Usually, a \ac{cfar} algorithm \cite{cfar} or a variation thereof defines the threshold. A clustering algorithm groups nearby detected signals, and the respective cluster means are input to the tracking stage. In this part of the pipeline, the track management determines whether to assign the measurement to a track, open a new track, discard the measurement or delete non-active tracks. Before updating the track, the measurement has to be filtered by the tracking filter based on the last position and an underlying movement model. The \ac{ukf} is a commonly used tracking filter \cite{ukalman}.
\begin{figure}[htbp]
\centerline{\includegraphics[width=\linewidth]{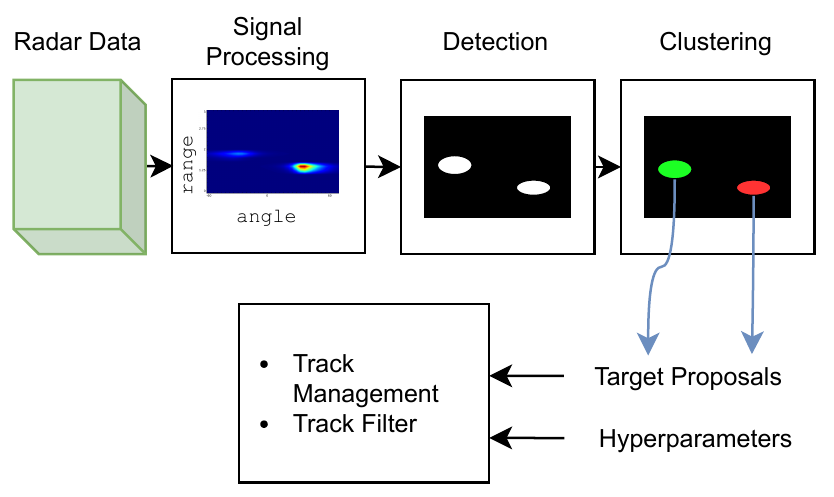}}
\caption{High-level description of a Radar Tracking Pipeline.}
\label{fig:track_pipe}
\end{figure}
The presented tracking pipeline heavily relies on hyperparameters. Namely, the tracking performance depends on the gating threshold for assigning tracks and the covariance matrix of the measurement and state transition models. Typically, those hyperparameters are determined by an expert user with recorded data and ground truth positions evaluating the Normalized Estimation Error Squared (NEES). However, this approach is unlikely to perform well once the radar is deployed in a different environment. Thus, recent work proposed to use \ac{rl} to tackle the combinatorial problem of finding the best set of parameters for any scenario \cite{hyperpar}.

\subsection{Reinforcement Learning} \label{subsec:RL}
In \ac{rl} the problem is formalized as a \ac{mdp} by $(S, A, R, p, \gamma)$, where $S$ is the state space as radar sensor input, $A$ is the action space defined as hyperparameters, $R$ is the reward as tracking performance shown in \cite{hyperpar}, $p_{\pi}$ is the unknown transition probability between states following policy $\pi$ and $\gamma$ is the discount factor. Let $\tau = (s_t, a_t, r_t, s_{t+1})$ define the transition from state $s$ at time step $t$ to the next state $s_{t+1}$ following action $a_t$ with reward $r_t$.
In traditional \ac{rl} the goal is to maximize the sum of expected rewards 
\begin{equation}
    \sum_t \mathbb{E}_{(s_t , a_t) \sim p_{\pi} [r(s_t, a_t)]}.
\end{equation}
This can be achieved by value iteration methods \cite{valueiter}. There, we define a Q-Value $Q(s_t,a_t)$ for each state and action pair that estimates the expected reward. Afterward, for each state we select the action which maximizes the Q-Value. With the advancements of neural networks in the context of universal function approximators \cite{lu2020universal}, methods like DQN \cite{dqn} use neural networks for the Q-value approximation.
However, value iteration methods become infeasible when the action space is continuous. Thus, the policy gradient theorem \cite{kakade2001natural} provides a way to update the policy for continuous actions. The combination of Q-Learning and policy gradients forms the basis of actor-critic methods. 
In recent years, policy gradient-based algorithms like Proximal Policy Optimization (PPO) \cite{schulman2017proximal} or actor-critic methods like Soft Actor-Critic (SAC) \cite{haarnoja2018soft} have shown great success. The particular choice of the algorithm depends on the task at hand. On-policy methods update the policy only with transitions from following the \textit{same} policy, which is data inefficient. In contrast, off-policy methods store transitions in a replay buffer and update the policy from these. The current policy and the policy that collected the transitions can be unrelated, which is the core concept of off-policy methods. Since the policy update is possible with transitions from any policy, the collected data is reusable. Due to the data efficiency, off-policy methods are widely used in applications where data is scarce \cite{hesterlearning}. Despite advancements in \ac{rl}, transferring \ac{rl} agents to real-world problems remains challenging \cite{dulac2021challenges}. To address these limitations, \cite{metarlapplica} has shown how meta-learning can be used to transfer an \ac{rl} agent to a real-world problem.

\subsection{Meta Reinforcement Learning} \label{subsec:MetaL}
State-of-the-art \ac{ml} models usually require many data \cite{silver2016mastering}. However, humans can transfer experiences to have a ``educated guess'' about new tasks based on their knowledge from related tasks, e.g., recognizing an animal from the zoo in the TV. Meta-learning has been introduced to support the training of \ac{ml} models to imitate this human capability. To this end, each task is associated with a respective dataset $\mathcal{D}$. The meta-learner aims to increase the performance on unseen test tasks, which are different from the tasks it has been trained on. This procedure can be adapted to many machine learning fields, such as supervised learning \cite{ren2018meta} and \ac{rl} \cite{metarl}. In \ac{mrl}, the task difference can be induced by different reward functions or environments, e.g., several Maze puzzle environments \cite{nam2022skill}. Successful algorithms as Pearl \cite{pearl}, Reptile \cite{reptile} or MAML \cite{maml} are designed to minimize the adaptation steps in new environments. As an example, adaptation is not feasible in a radar tracking application during inference due to the lack of rewards. Thus, in applications without reward signal during inference, domain generalization is key. Inspired by \cite{domain}, where the target objective of MAML is adapted to improve domain generalization, we adapt the design of Pearl to an efficient context variable computation.
In Pearl, these context variables are computed by a neural network from stored transitions of the test task. As a consequence, there is a need for context variables that can be efficiently computed from the input data and do not require reward information during inference. Although, the discussed methods increase the generalization, they do not guarantee the robustness in case the inference distribution differs from the training distribution.

\subsection{Uncertainty-based Reinforcement Learning} \label{subsec:UncRL}
In order to increase the robustness, highly uncertain predictions need to be classified as \ac{ood} in safety-critical applications, as shown in \cite{pacheco2020out}. At the same time, the radar tracker is often limited by the resolution of the radar settings. Thus, there is a limit on the amount of people who can be tracked simultaneously in the scene. In addition, moving disturbances like curtains can be detected as false targets. Consequently, the algorithm needs to detect those scenarios and notify the user that tracking is impossible. Thus, we have a bayesian setting, where we estimate the posterior probability $p(\theta| x)$ on the network parameters $\theta$ given the data $X$. The posterior is proportional to a prior believe on the distributional function $p(\theta)$ and the likelihood $p(x |\theta)$. Further, the posterior probability determines the uncertainty on the prediction. \newline
\ac{rl} literature proposed several ways to estimate the uncertainty on the neural network parameters:  dropout \cite{gal2016dropout} or bootstrap DQN \cite{osband2016deep} and bootstrap DQN with random prior functions \cite{osband2018randomized}. Dropout applies a random Bernoulli mask on the neural network weights to prevent co-adaptation. In \cite{gal2016dropout}, they argue that the dropout rate distribution approximates the posterior distribution.
However, dropout is insufficient as a posterior estimation since the dropout rate is not dependent on the data. Hence, this method can not differentiate between data points it has seen once or multiple times, as pointed out in \cite{osband2018randomized}. A more prominent way is the bootstrap method with random prior functions. The bootstrap DQN defines a neural network with $K$ heads that estimate $K$ Q-Values $Q_{i=1,...,K}(s_t, a_t)$. The update is computed for every single head with the Temporal Difference (TD) error, shown in Equation \ref{eq:td}, with the reward $r_t$, discount factor $\gamma$, state $s$, action $a$ and time step $t$.
\begin{equation}
    \delta_i = Q_i(s_t,a_t) - (r_t + \gamma \max_{a} Q_i(s_{t+1}, a))
    \label{eq:td}
\end{equation}
For each state and action, the variation in the head prediction estimates the uncertainty about the given scene. In order to enhance this variation, a binary mask selects the active heads during training, such that the heads observe different data. In addition, randomized prior functions dictate the network behavior wherever is no training data. The random prior function can be estimated by a non-trainable, random initialized neural network that processes the input data \cite{osband2018randomized}. \newline
Since radar tracking is a continuous action space problem, bootstrap DQN is not applicable. However, DQN is a critic-only method, referring to \cite{konda1999actor}. Hence, we can adapt the bootstrap mechanism with priors to actor-critic methods without loss of generality.

\section{Approach} \label{sec:approach}
In this section we present an approach for robust radar tracking with meta-\ac{rl} in combination with \ac{ood} detection. In Section \ref{subsec:radar_signal}, we introduce the signal processing of the radar-based tracking chain and \ac{rai} generation. Afterward, we describe context priors in Section \ref{subsec:rand_prios} and how they encode the complexity of the ongoing \ac{rl} task. Furthermore, as described in Section \ref{subsec:meta_rl}, the information is used for meta-\ac{rl}, which bears the generalization on tasks of different complexity. Finally, the approach is completed by Section \ref{subsec:ood_det}, where, using uncertainty, the approach can detect \ac{ood} tasks where the tracking is likely to fail.

\subsection{Radar Signal Processing}\label{subsec:radar_signal}
The received data at each time step is a three-dimensional array of shape $(N_C, N_S, N_{rx})$, dependent on the number of chirps $N_C$, the number of samples $N_S$ per chirp, and the number of receiving antennas $N_{rx}$. Let the axis along the chirps denotes the slow time, and the axis along the number of samples the fast time. We subtract the mean in the fast time to prevent any transmit/receiving antenna leakage. Additionally, we subtract the mean in the slow-time as a high-pass filter to remove completely static target information, e.g., furniture. As the last step, we transform the array to the frequency domain with a \ac{fft}. The range corresponds to frequency along the fast time. Let $B$ be the bandwidth of the transmit signal, $T_c$ the active chirp time, $f_s$ the sampling frequency, and $c_0$ the speed of light. The maximum range is defined in Equation \ref{eq:max_range}. 
\begin{equation}
    R_{max} = \frac{f_s}{2} \frac{c_0 T_c}{B}
    \label{eq:max_range}
\end{equation}
The \ac{aoa} can be estimated by the phase difference between the minimum two receiving antennas using Digital Beamforming (DBF), which depends on the antenna gain at a given \ac{aoa}. Here we use the Capon beamformer described in \cite{capon}.
This paper uses a \ac{fmcw} radar with one transmit and three receiving antennas in a triangular alignment. Hence two antennas are used for the azimuth \ac{aoa} estimation. By utilizing a bandwidth of 1 GHz, chirp time of 399 $\mu s$, and the sampling frequency of 2 MHz, the maximum detection range is approximately $5 m$. As example, we show an \ac{rai} with two targets in Figure \ref{fig:radar_rai}. In our method, the \ac{rai} is used as input to the meta-\ac{rl} algorithm.
\begin{figure}[htbp]
\centerline{\includegraphics[width=\linewidth]{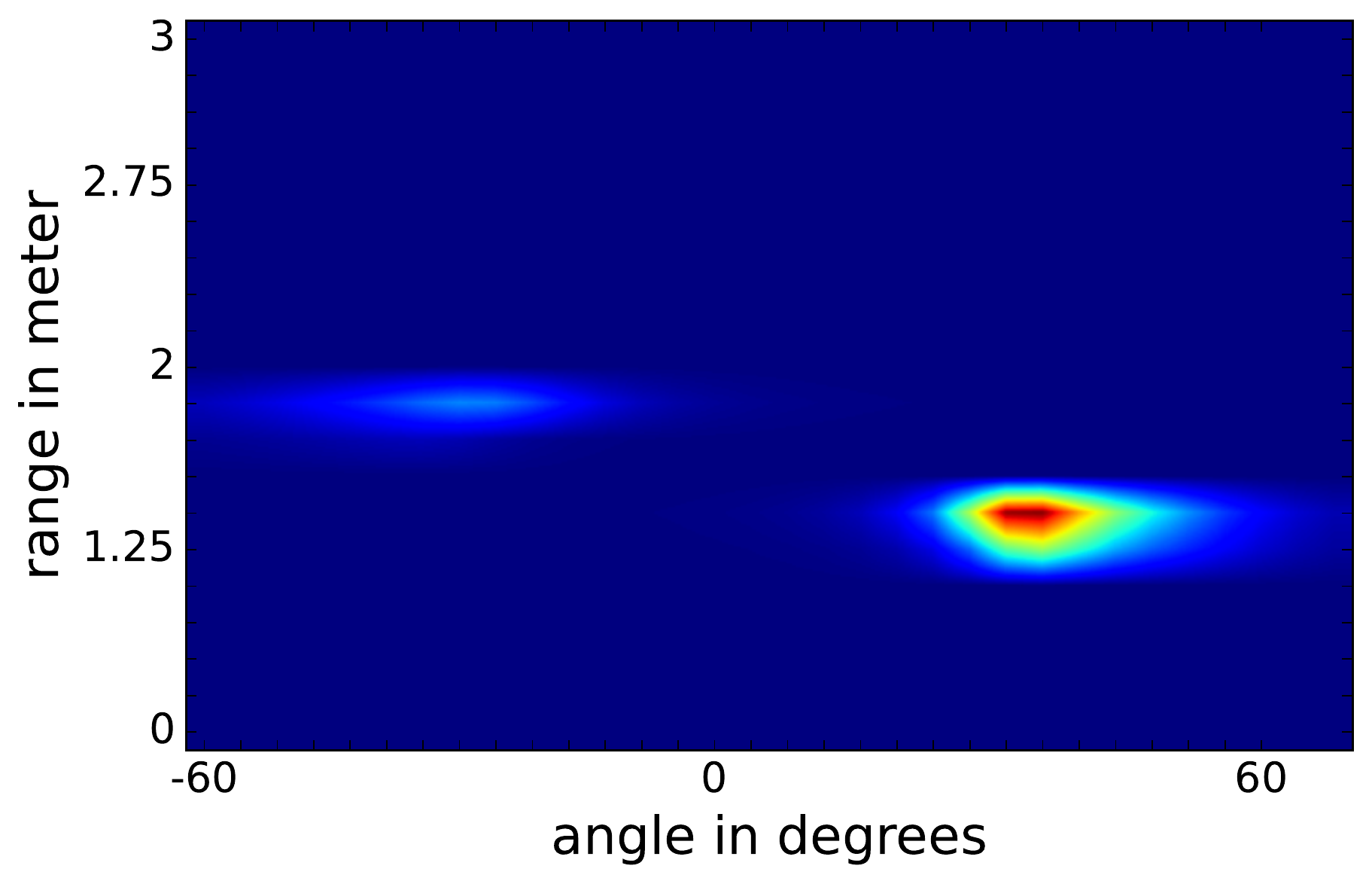}}
\caption{Example of a Range-Angle Image with two targets. The color encodes energy intensity from high (\textcolor{red}{red}) to low (\textcolor{blue}{blue}). }
\label{fig:radar_rai}
\end{figure}

\subsection{Random Priors and Context Variables} \label{subsec:rand_prios}
In Section \ref{sec:back} we introduced random priors for uncertainty estimation and context variables for meta-learning. Both are random vectors computed from the input data and serve as additional input to the neural network. However, random priors are computed by a non-trainable and random initialized neural network. In contrast, context variables in \cite{pearl} are Gaussian random variables computed from task-specific transitions following the context-conditioned policy. In that way, the context variable infers information about the task difficulty from the reward. For radar tracking, the reward is given by the tracking performance, which requires ground truth positions. This information is not available during inference time. However, the average intensity and the variance in the \ac{rai} are increasing with the number of people in the current scene, as shown in Figure \ref{fig:radar_mean} and Figure \ref{fig:radar_std}.
Since the \ac{rai} distribution is proportional to the intensity of the tracking targets, we can use the \ac{rai} distribution to encode the task difficulty. To this end we use the mean $\mu_{rai}$ and standard deviation $\sigma_{rai}$ of the radar input to define a Gaussian context prior $\mathcal{N}(\mu_{rai}, \sigma_{rai})$.

\begin{figure}
     \centering
     \begin{subfigure}[b]{0.4\textwidth}
         \centering
         \includegraphics[width=\textwidth]{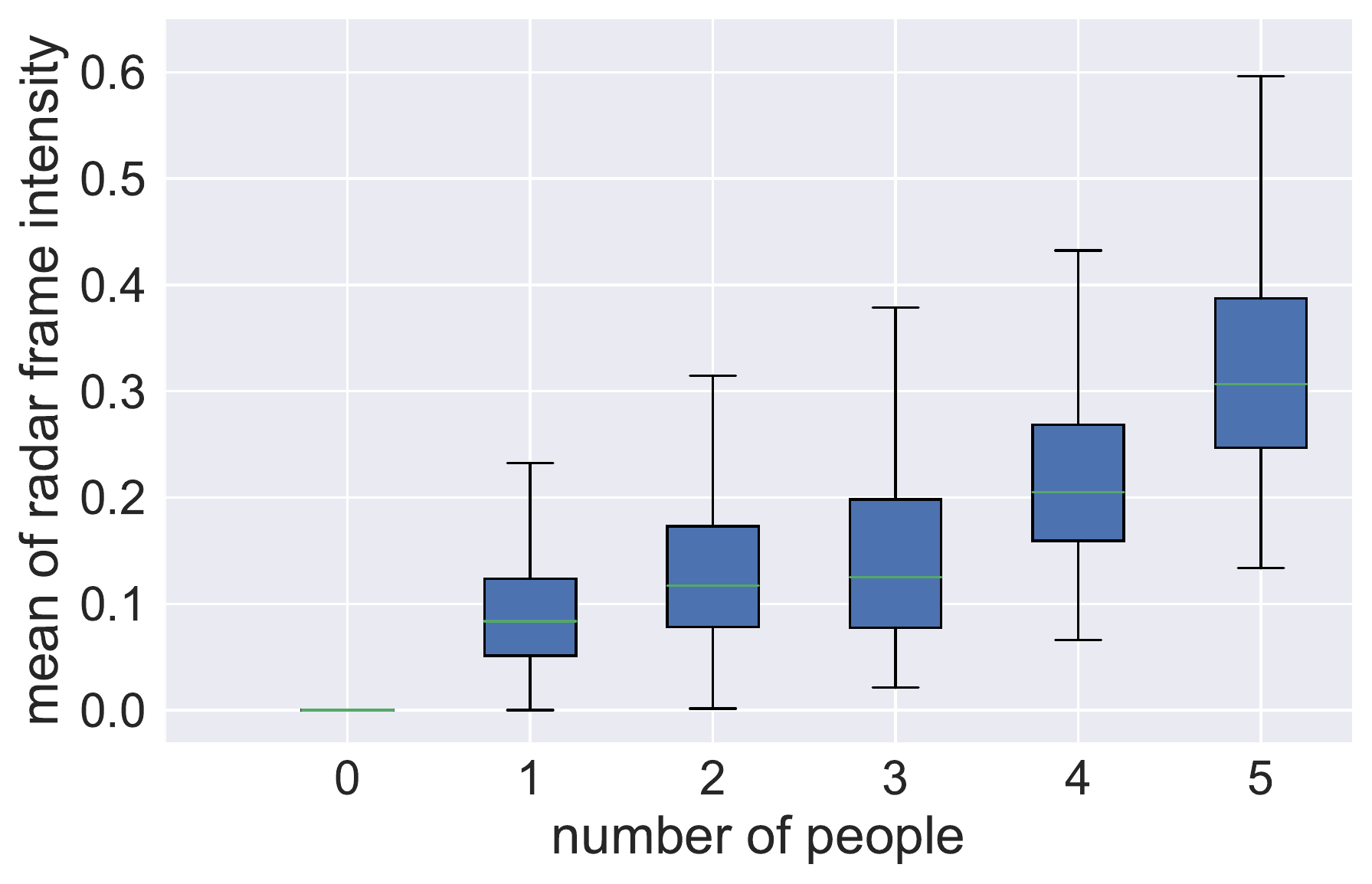}
         \caption{Distribution of mean Range-Angle Image intensity values with respect to the number of targets.}
         \label{fig:radar_mean}
     \end{subfigure}
     \hfill
     \begin{subfigure}[b]{0.4\textwidth}
         \centering
         \includegraphics[width=\textwidth]{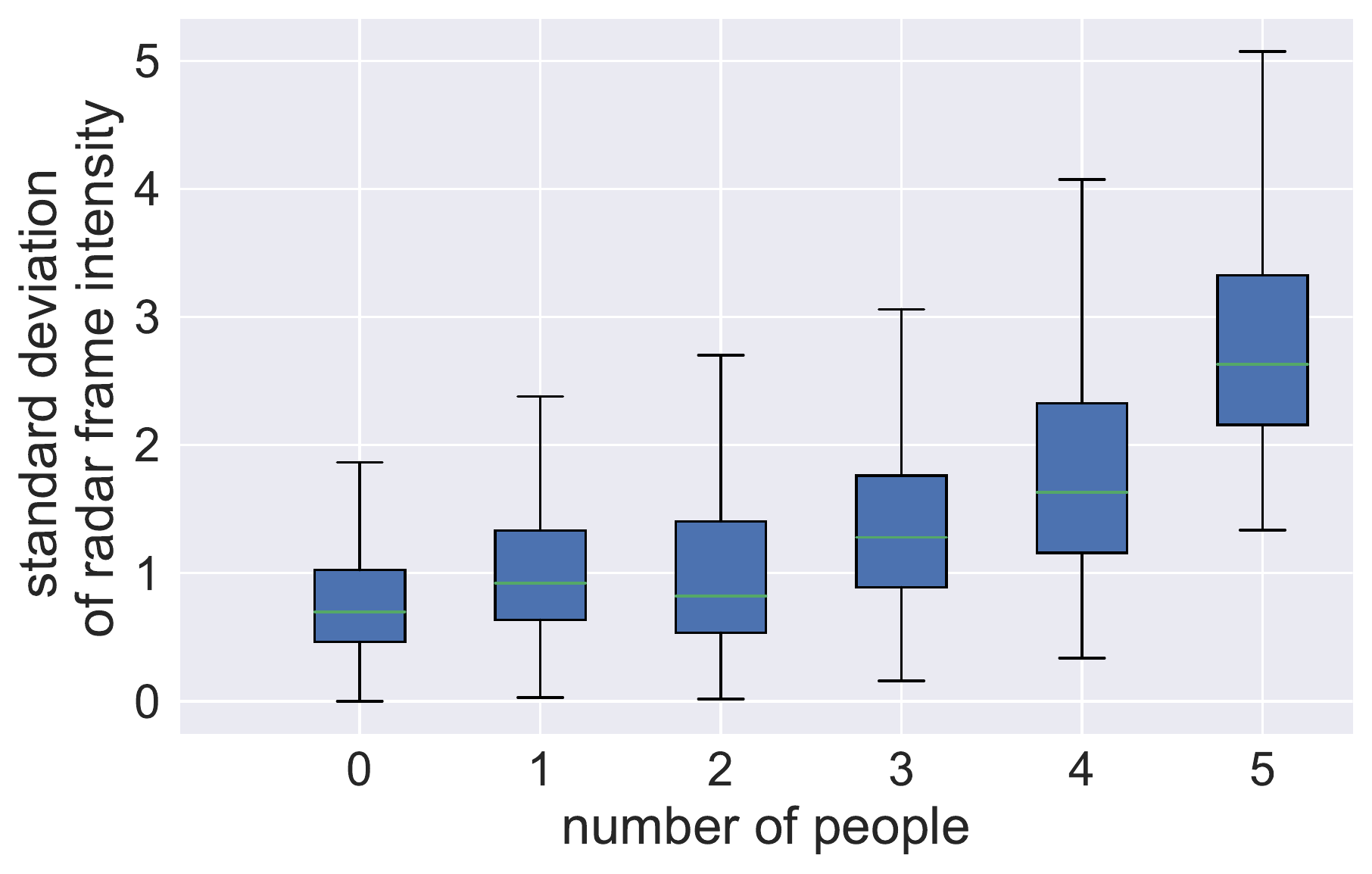}
         \caption{Standard deviation distribution of Range-Angle Image intensity values with respect to the number of targets.}
         \label{fig:radar_std}
     \end{subfigure}
 \caption{Comparison of \ac{rai} intensity distributions with respect to the number of targets.}
\label{fig:radar_dists}
\end{figure}

\subsection{ Meta-RL with Context Prior} \label{subsec:meta_rl}
In this \ac{mrl} problem, we define different rooms as tasks since the goal is to optimize the tracking parameters independent of the room. In addition, we divide the different rooms into training and test tasks. The test tasks are not observed during training. Moreover, we report the evaluation reward as the average reward over the test tasks, while the \ac{mrl} algorithm is trained on the train tasks. A detailed description of the training and testing procedure is given in Figure \ref{fig:metarl_pip}.
\begin{figure}[htbp]
\centerline{\includegraphics[height=0.8\linewidth]{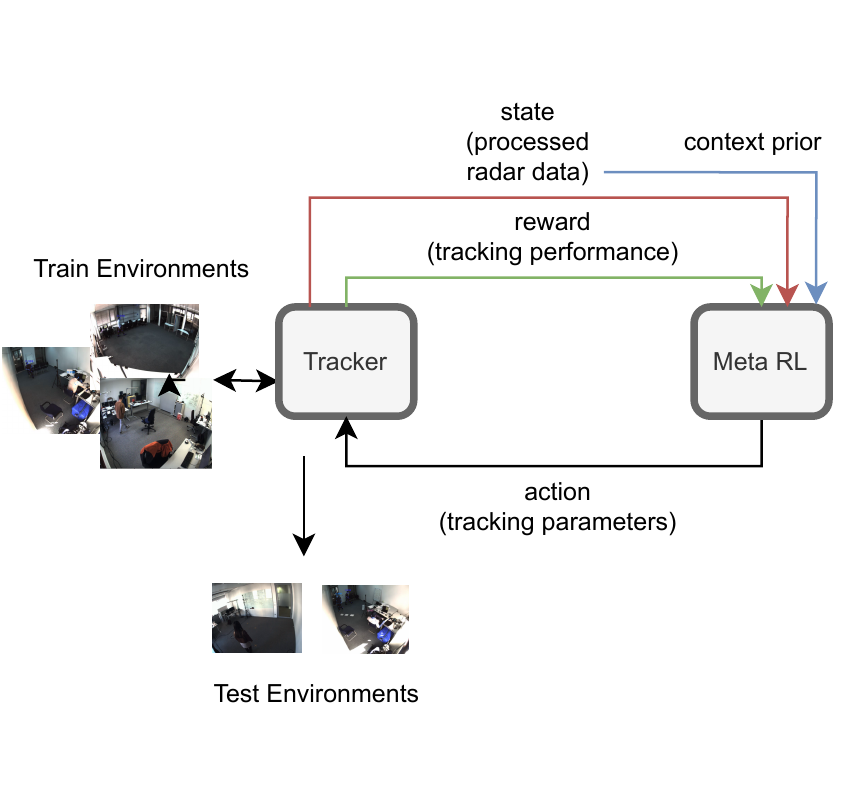}}
\caption{Overview of the meta-training and evaluation procedure.}
\label{fig:metarl_pip}
\vskip -15pt
\end{figure}
The reward is given in Equation \ref{eq:reward}, where $\hat{N}$ is the predicted number of targets, $N$ is the true number of targets, $M=\min(\hat{N}, N)$, $\hat{\mu_k}$ is the predicted mean, $\hat{\Sigma_k}$ the predicted covariance and $p_k \sim \mathcal{N}(\hat{\mu}, \hat{\Sigma})$, as described in \cite{hyperpar}.
\begin{equation}
    -R_t  = \rho(\hat{N_t}, N_t) + \frac{1}{M} \sum_{k=0}^{M}1- p_k(P_k) 
    \label{eq:reward}
\end{equation}
The first term $\rho$ describes the relative error between the predicted and actual number of tracks. The second term evaluates the likelihood of missing the ground truth positions incorporating the variance of the \ac{ukf}. For stability reasons, $p(\cdot)$ is clipped to $1$. \newline
As baseline \ac{rl} algorithm we use SAC, a state-of-the-art off-policy actor-critic method from \cite{haarnoja2018soft}. \newline
By design of the context prior, gradient calculation can be omitted and we can use the loss functions for the actor from \cite{haarnoja2018soft}, given in Equation \ref{eq:policy_loss} and the critic loss for the individual bootstrap heads in Equation \ref{eq:critic_loss}.

\begin{equation}
    J_{\pi}(\phi) = \mathbb{E}_{s_t\sim D} \left [ D_{KL} \left( \pi (\cdot | \mathbf{s}_t) \bigg \|  \frac{exp(Q_{\theta}(\mathbf{s}_t, \cdot)}{Z_{\theta}(\mathbf{s}_t)} \right) \right]
    \label{eq:policy_loss}
\end{equation}

\begin{equation}
    J_{Q}^i(\theta) = \mathbb{E}_{(\mathbf{s}, \mathbf{a}, \mathbf{r}, \mathbf{s`}) \sim D } \left [ Q^i_{\theta}(\mathbf{s}, \mathbf{a}) - (\mathbf{r}(\mathbf{s}, \mathbf{a}) + \bar{Q}^i(\mathbf{s'} \mathbf{a}) ) \right]^2
    \label{eq:critic_loss}
\end{equation}

Furthermore, the critic uses the bootstrap mechanism. To this end, we define a Base Network that computes an embedding vector $x$ from the state action pairs. During training, we compute a binary mask in every epoch, determining whether the parameters for specific heads are updated. For every active head, we sample a context vector from the context prior distribution and add it to $x$. Afterward, every head predicts a Q-Value $Q^i(s, a)$. In detail, we describe the bootstrap critic algorithm in Algorithm \ref{algo:boot_critic}. We apply the exact prior mechanism for the actor network that predicts the actions from the current state. A detailed description of the used networks is given in Figure \ref{fig:networks}. \newline
For meta-learning, we update the parameters with the accumulated losses from every training task, as shown in \cite{pearl}.

\begin{algorithm}
\caption{Bootstrap Critic with Context Prior}
\label{algo:boot_critic}
\begin{algorithmic}[1]
\REQUIRE Bootstrap heads $\{H^i\}_{i=1,...N}$, Buffer $B$, learning rate $\alpha$

  \STATE Initialize $BaseNetwork$ parameters $\theta$
  \STATE Initialize Head parameters $\delta_i$
  \WHILE{ not done}
  \STATE compute active head mask: $m_i \sim \mathcal{B}(0,1)$
  \STATE active Heads: $H_m^i$
  \STATE $(s_t, a_t, r_t, s_{t+1})$ from $B$
  \STATE $\mu = mean(state)$
  \STATE $\sigma = std(state)$
  \STATE context prior: $\mathcal{N}(\mu, \sigma)$
  \FOR{head in $H_m^i$}
  \STATE $x = BaseNetwork(s_t, a_t)$
  \STATE $z_i \sim \mathcal{N}(\mu, \sigma)$
  \STATE $Q_i = head(x + z_i)$
  \STATE $\delta_i = \delta_i - \alpha \nabla_{\delta_i} J_{Q}^i$
  \ENDFOR
  \STATE $\theta = \theta - \alpha  \frac{1}{M}\sum_i^M \nabla_{\theta} J_{Q}^i$
  \ENDWHILE
\end{algorithmic} 
\end{algorithm}

\begin{figure*}
     \centering
     \begin{subfigure}[t]{0.28\textwidth}
     \vskip 0pt
         \centering
         \includegraphics[width=\textwidth]{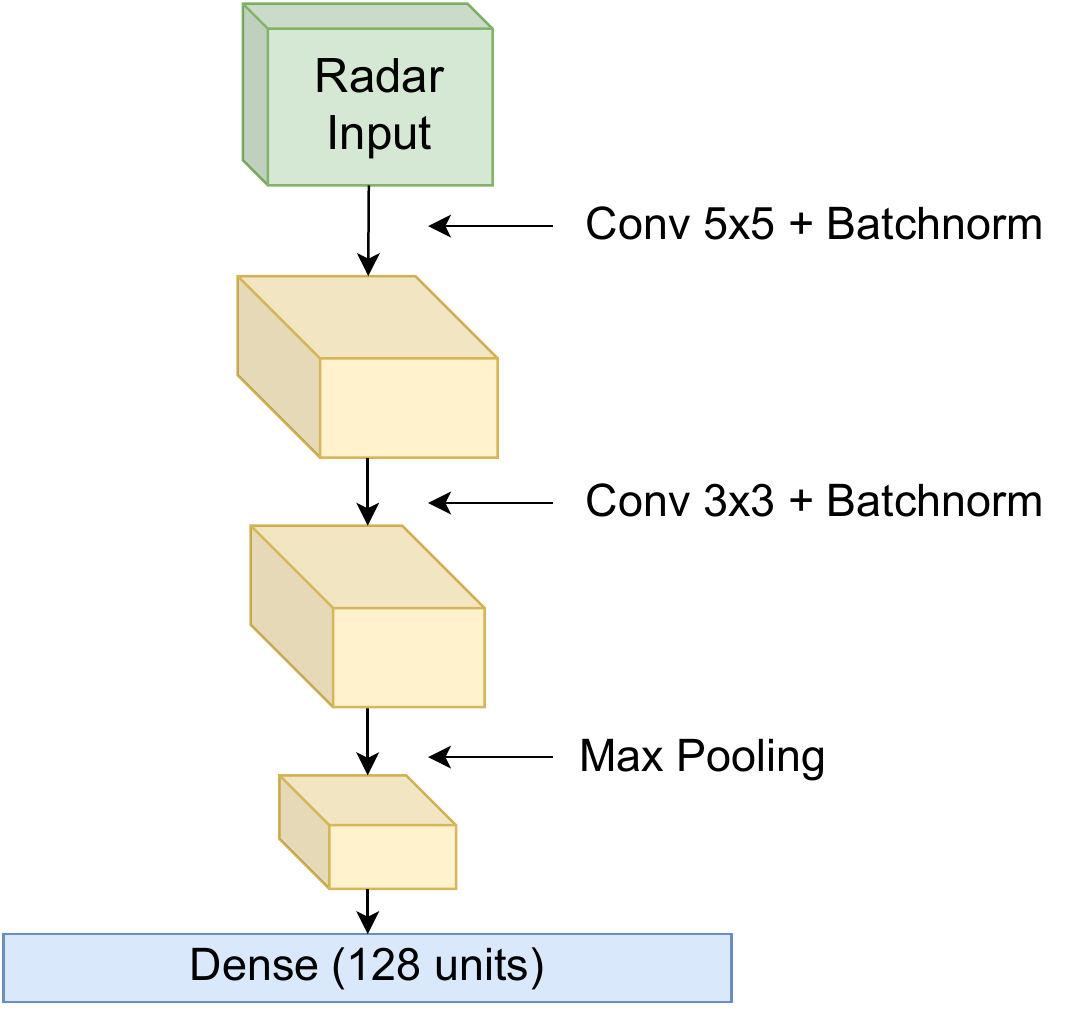}
         \vskip 19 pt
         \caption{Base Network}
         \label{fig:base_net}
     \end{subfigure}
     \hfill
     \begin{subfigure}[t]{0.3\textwidth}
     \vskip 0pt
         \centering
         \includegraphics[width=\textwidth]{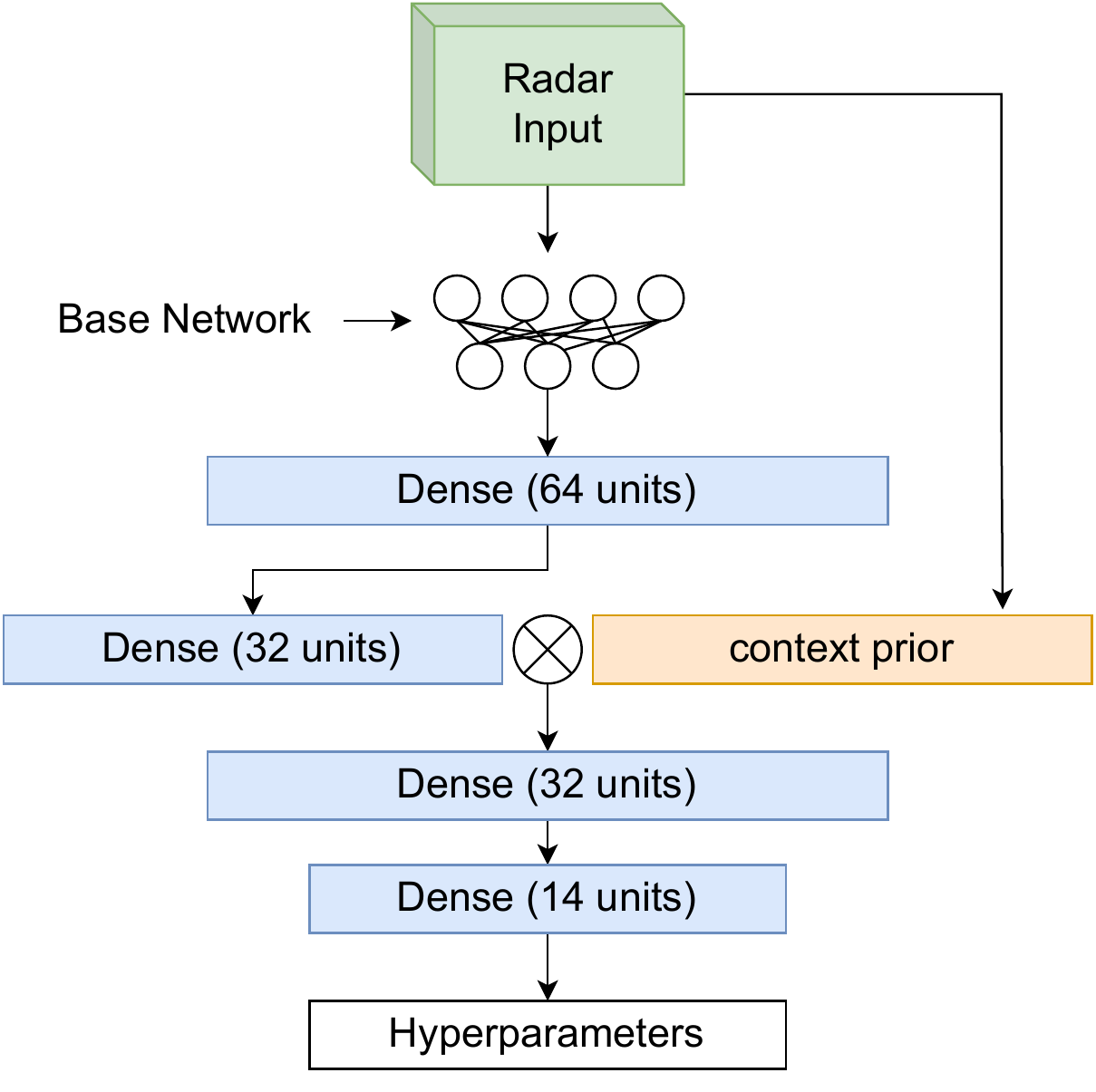}
         \caption{Actor Network}
         \label{fig:actor_net}
     \end{subfigure}
     \hfill
     \begin{subfigure}[t]{0.32\textwidth}
     \vskip 0pt
         \centering
         \includegraphics[width=\textwidth]{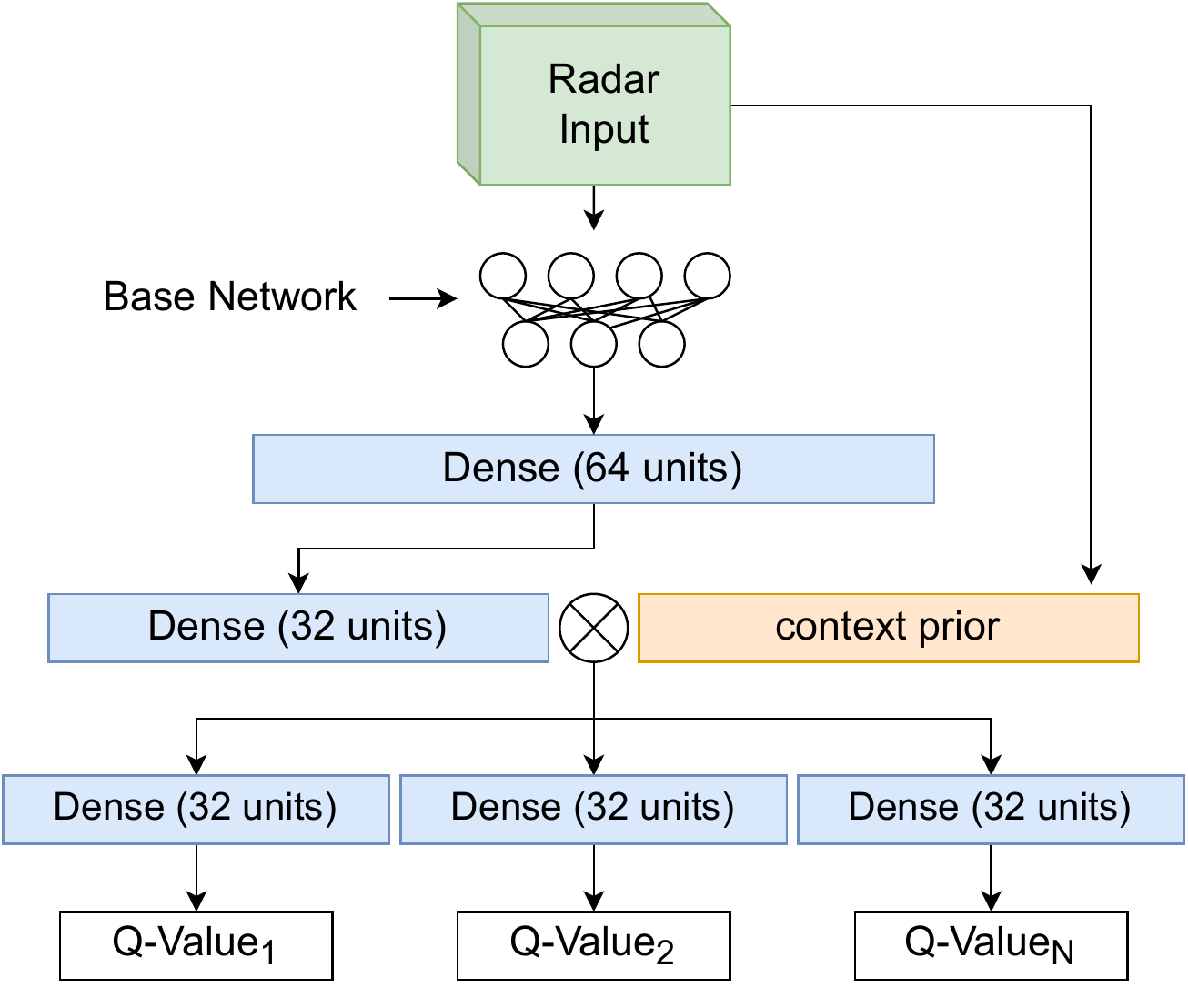}
         \vskip 13 pt
         \caption{Multi-Head Critic Network}
         \label{fig:critic_net}
     \end{subfigure}
 \caption{Neural Network designs for actor-critic \ac{rl} with context prior}
 \label{fig:networks}
\end{figure*}

\begin{algorithm}
\caption{Meta-Training}
\label{algo:metarl}
\begin{algorithmic}[1]
 \REQUIRE training tasks $ \{T^{i}\}_{i=1,...,T}$
 \STATE Initialize parameters $\theta$, $\phi$
  \STATE Initialize buffer for each task $B_i$
  \STATE learning rates $\alpha_1$ , $\alpha_2$
  \WHILE{ not done}
  \FOR{task in $T^i$}
  \STATE rollout $\pi(\theta|s)$ and store transitions in $B_i$
  \STATE sample batch $b^i \sim B_i$
  \STATE compute $\mathcal{L}_{boot-critic}^i(b^i)$
  \STATE compute $\mathcal{L}_{actor}^i(b^i)$
  \ENDFOR
  \STATE $\theta  = \theta - \alpha_1 \nabla_{\theta}\sum_{i=1}^N \mathcal{L}_{boot-critic}^i$
  \STATE $\phi  = \phi - \alpha_2 \nabla_{\phi}\sum_{i=1}^N \mathcal{L}_{actor}^i$
  \ENDWHILE
\end{algorithmic} 
\end{algorithm}

\subsection{Out-of Distribution Detection}\label{subsec:ood_det}
In literature, \ac{ood} approaches aim to detect whether an environment has not been seen yet. The challenge in our setup is, for new scenarios, to generalize well and detect low-performance simultaneously. \newline
We apply the bootstrap mechanism for the critic for \ac{ood}, as shown in \cite{oodrl}. The critic aims to predict the future reward, and the variation along multiple predictions is a measure of uncertainty. As we see from Figure \ref{fig:radar_dists}, a huge variance in energy peaks between two detected targets in a radar image, is an example of increased task complexity. Thus, our proposed context prior emphasizes higher variation for a more difficult task. The training procedure of the bootstrap critic is shown in Algorithm \ref{algo:boot_critic}. 
In order to detect \ac{ood} samples during inference, we define a threshold $c$ as given in Equation \ref{eq:ood_t}, 
\begin{equation}
    c =  \mu_{head} + \alpha \sigma_{head}
    \label{eq:ood_t}
\end{equation}
where $\mu_{head}$ is the prediction mean of all heads, $\sigma_{head}$ is the standard deviation of the predictions and $\alpha$ is a hyperparameter that determines the size of the \ac{ood} detection interval.

\section{Results and Discussion} \label{sec:eval}
In this section, we present the detailed implementation settings for the experiments, starting with the used hardware and software tools, followed by the dataset specifications. The results for \ac{mrl} are presented in Section \ref{subsec:meta_results} and the results for \ac{ood} detection are shown in Section \ref{subsec:ood_results}.

\subsection{Implementation Settings and Dataset}
In the implementation, we used Tensorflow\textsuperscript{\texttrademark}- GPU v2.9.0 with CUDA\textsuperscript{\textregistered} Toolkit v11.2.0 and cuDNN v8.1.0. As a processing unit, we used the Nvidia\textsuperscript{\textregistered} Tesla\textsuperscript{\textregistered} P40 GPU, Intel\textsuperscript{\textregistered} Core i9-9900K CPU, and DIMM 16GB DDR4-3200 module of RAM. As a radar sensor for data recordings, we used three of Infineon Technologies XENSIV\textsuperscript{\texttrademark} 60 GHz BGT60TR13C \ac{fmcw} radar and three cameras for labeling purposes. The recorded dataset includes 200.000 radar frames from five different rooms, including human activities from zero to five people, divided into recordings of 350 frames. The radar data has been recorded with a frame rate of 10 Hz. We use Yolo-v5 \footnote{\url{https://github.com/ultralytics/yolov5}}, an object detection framework from Ultralytics\textsuperscript{\textregistered}, with the three cameras to gather ground truth positions for the recordings.\newline
For meta-training, we take recordings from three rooms and use the other two rooms for the evaluation. The number of frames per environment is evenly split for the number of targets.

\subsection{Domain Generalization}\label{subsec:meta_results}
The current policy obtains the \ac{mrl} results in two unknown test environments. We report the average evaluation reward over the test environments after each iteration. The optimal reward is 0, and a higher reward indicates better tracking performance. The reward formulation is given in Equation \ref{eq:reward} and leverages the distance between predictions and ground truth and the number of false targets.
As a baseline, we use the performance of the tracker using fixed hyperparameters determined by an expert user. In addition, we compare our method against MAML with the formulation from \cite{domain} and Reptile.
In Figure \ref{fig:meta_results}, we show that our proposed method improves $35 \%$ over the baseline and $16 \%$ over the comparable \ac{mrl} methods. In addition, our proposed method explores more efficiently and is improving while MAML and Reptile saturate.
\begin{figure}[htbp]
\centerline{\includegraphics[width=\linewidth]{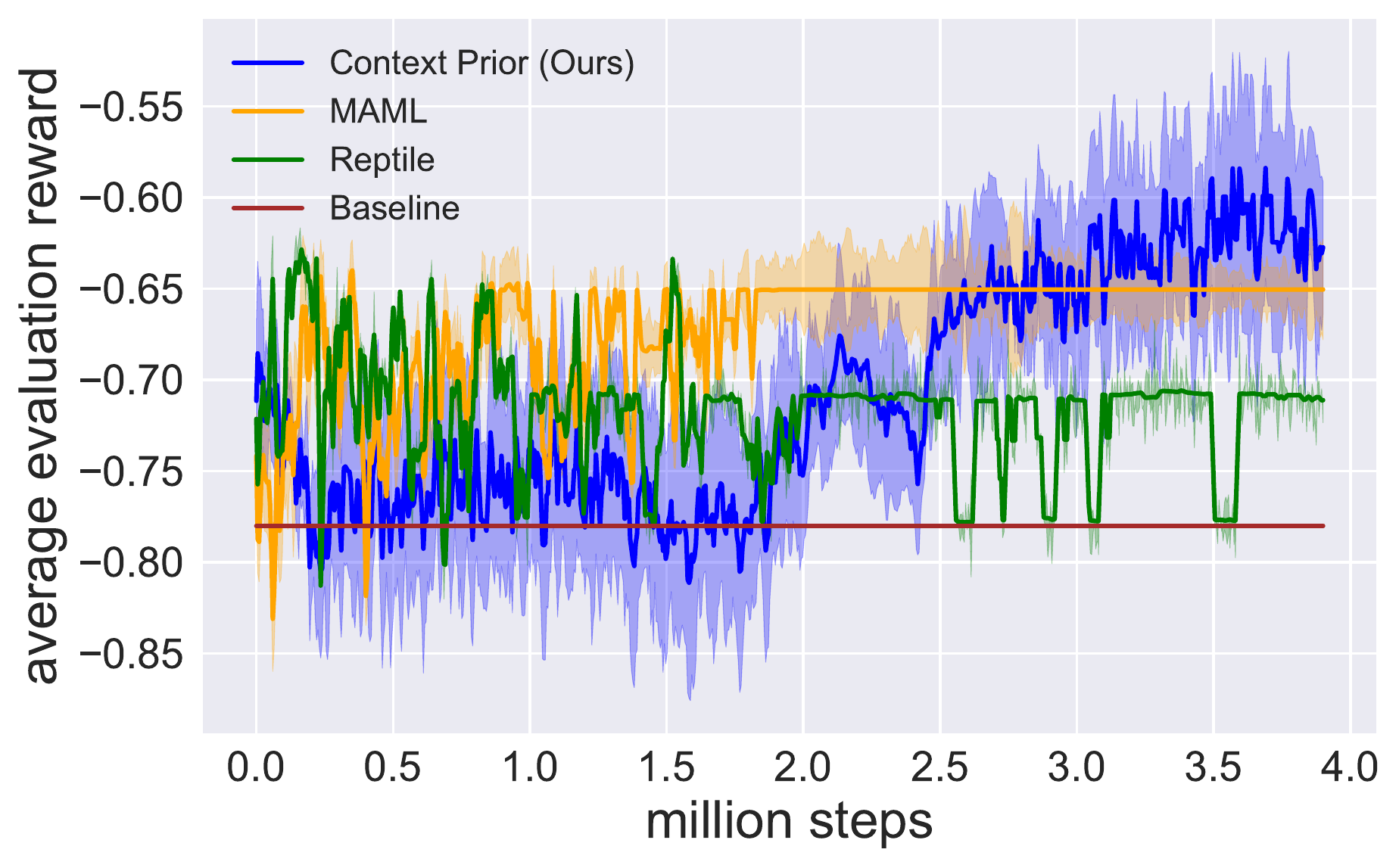}}
\caption{Comparison of the context prior \textcolor{blue}{(blue)} against MAML \textcolor{orange}{(orange)} and Reptile \textcolor{Green}{(green)}. The context prior reaches higher peak performance while MAML and Reptile saturate to almost static parameters.}
\label{fig:meta_results}
\end{figure}

The effect of adaptive parameter selection, where the reward focuses on the distance and false targets, is depicted in Figure \ref{fig:track_results}. There we see that our approach can estimate the correct number of tracks and is more accurate with a \ac{rmse} of $1.45$ compared to the baseline (\ac{rmse} of $1.94$) in estimating the target's angular position (x-position).
\begin{figure}[htbp]
\centerline{\includegraphics[width=\linewidth]{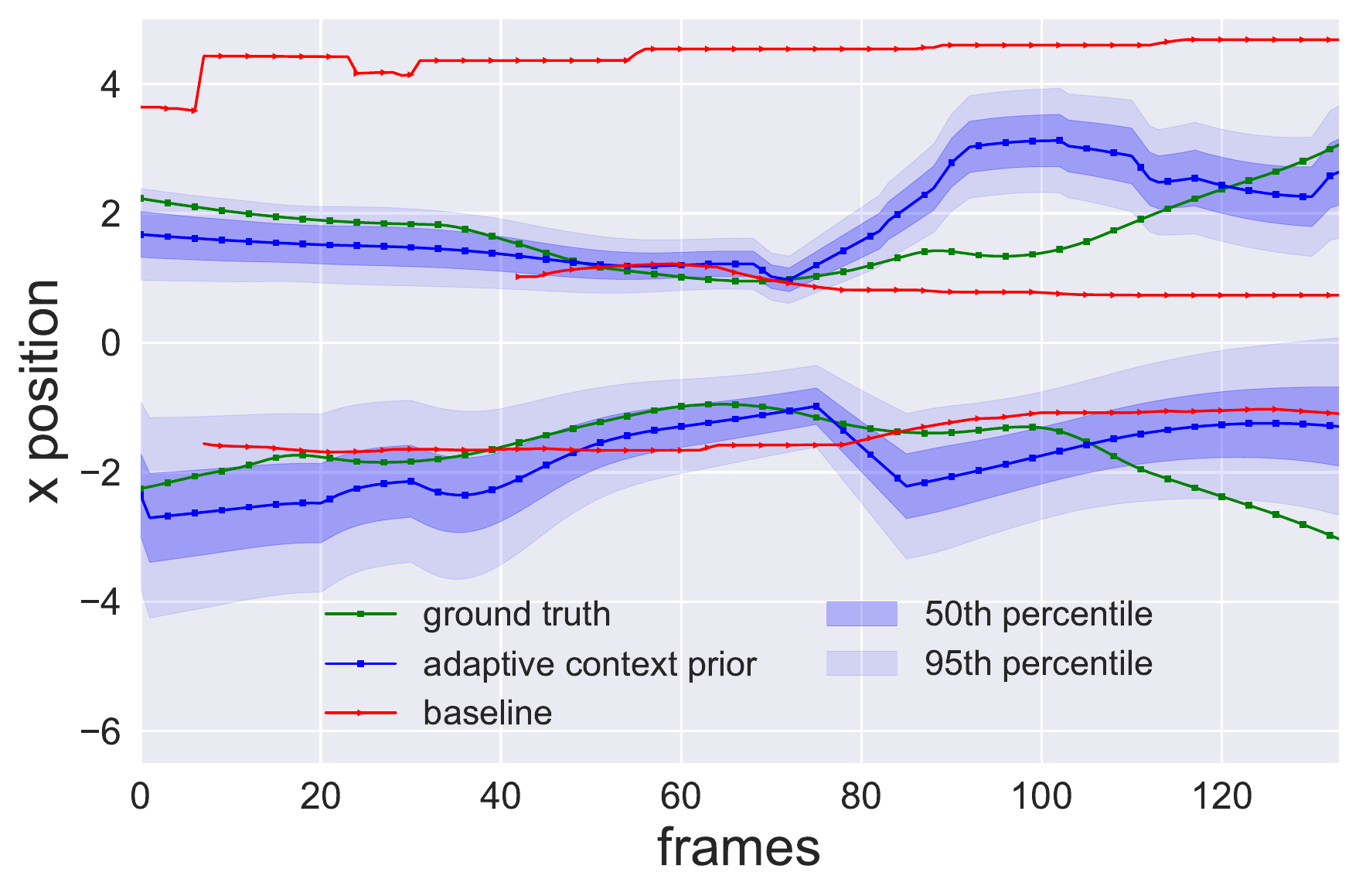}}
\caption{Comparison of our adaptive approach with the baseline for estimating the angular direction (x-position in Cartesian coordinates). The adaptive approach predicts the correct number of tracks while resulting more accurate in estimating the position with an RMSE of $1.45$ incorporating the predicted variance of the \ac{ukf} prediction.}
\label{fig:track_results}
\end{figure}
To further gain insights on meta-learning, we scale the reward during training to observe the impact on the environment generalization. Since the reward signal is used to improve tracking, and the context prior is used for environment generalization, weighting the reward higher than the context prior should decrease the generalization capabilities. We report the peak performance of the \ac{mrl} approach in the test environments with different reward scaling factors during training n Table \ref{tab:ablation_meta}. There we show that scaling the reward by two results in the highest average evaluation reward. Further we see that scaling the reward far off the context prior leads to worse evaluation performance, which supports our hypothesis.

\begin{table}[t]
    \centering
    \begin{tabular}{c|c}
         Reward scale factor & Best Reward \\ \hline
         1 & -0.58 \\
         \textbf{2} & \textbf{-0.51} \\
         5 & -0.61 \\
         10 & -0.68 
    \end{tabular}
    \caption{Study about the effect of reward scaling on the average test evaluation reward.}
    \label{tab:ablation_meta}
\end{table}

\subsection{Uncertainty-based OOD Detection}\label{subsec:ood_results}
With our defined setup, the expected limit of the tracking system is up to three targets. Hence, we aim to predict scenes with more than three targets as \ac{ood}. In Figure \ref{fig:critic_dists} we show the distributions of the critic predictions. Both mean and standard deviation are proportional to the number of people in the scene. \newline
Since we have fewer \ac{ood} scenarios (four and five targets) than training scenarios (zero to three targets), we report the F1-Score, which considers data imbalance.
With the threshold defined in Equation \ref{eq:ood_t}, $\alpha$ determines the size of the detection interval. Thus, choosing $\alpha$ very small focuses on accurately predicting all the \ac{ood} scenarios and vice versa, leveraging precision and recall. With $\alpha=0.17$ we reach a maximum F1-Score of $72\%$. 
\begin{figure}
     \centering
     \begin{subfigure}[b]{0.4\textwidth}
         \centering
         \includegraphics[width=\textwidth]{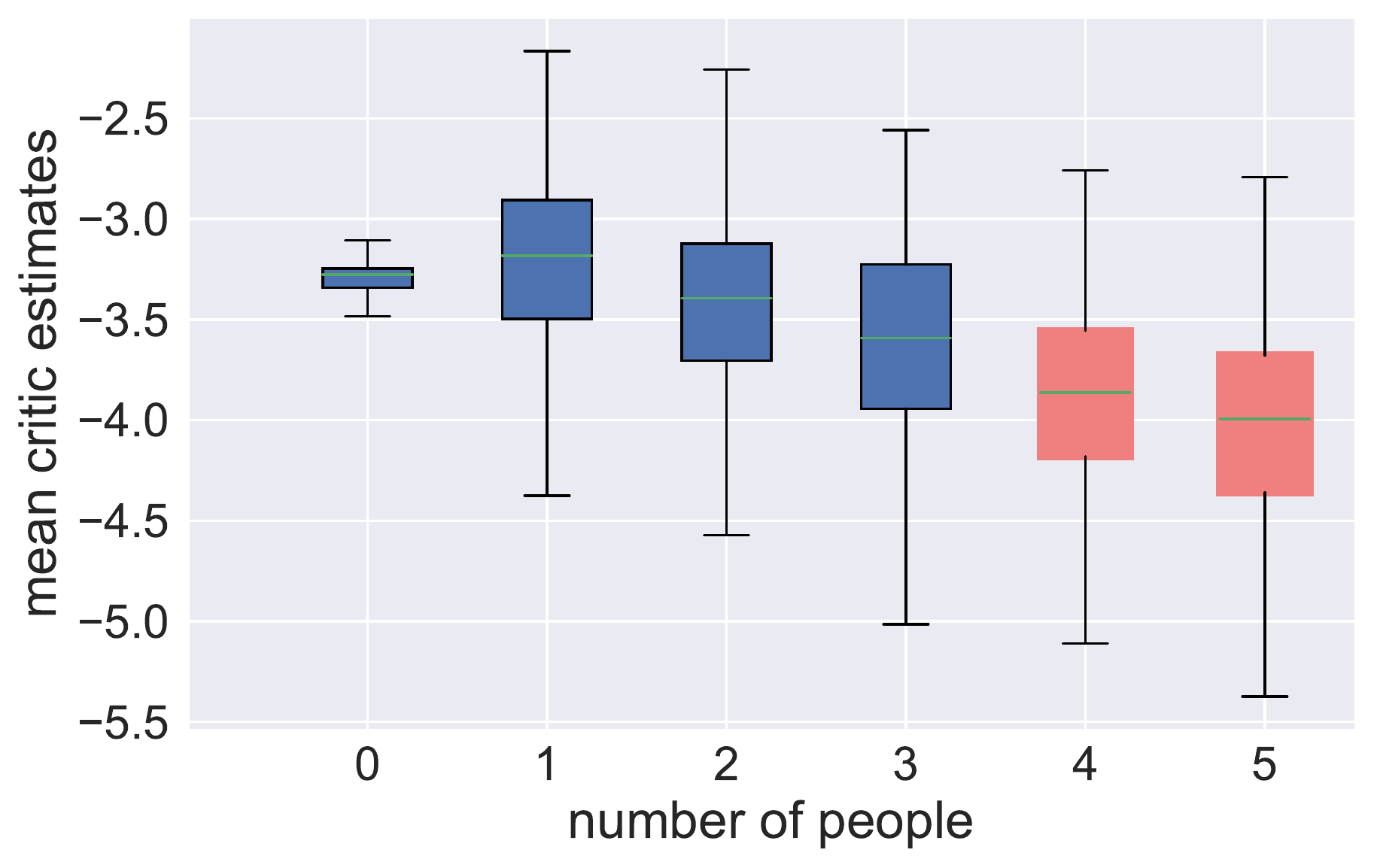}
         \caption{Mean Distribution of critic estimates on test environments.}
         \label{fig:critic_mean}
     \end{subfigure}
     \hfill
     \begin{subfigure}[b]{0.4\textwidth}
         \centering
         \includegraphics[width=\textwidth]{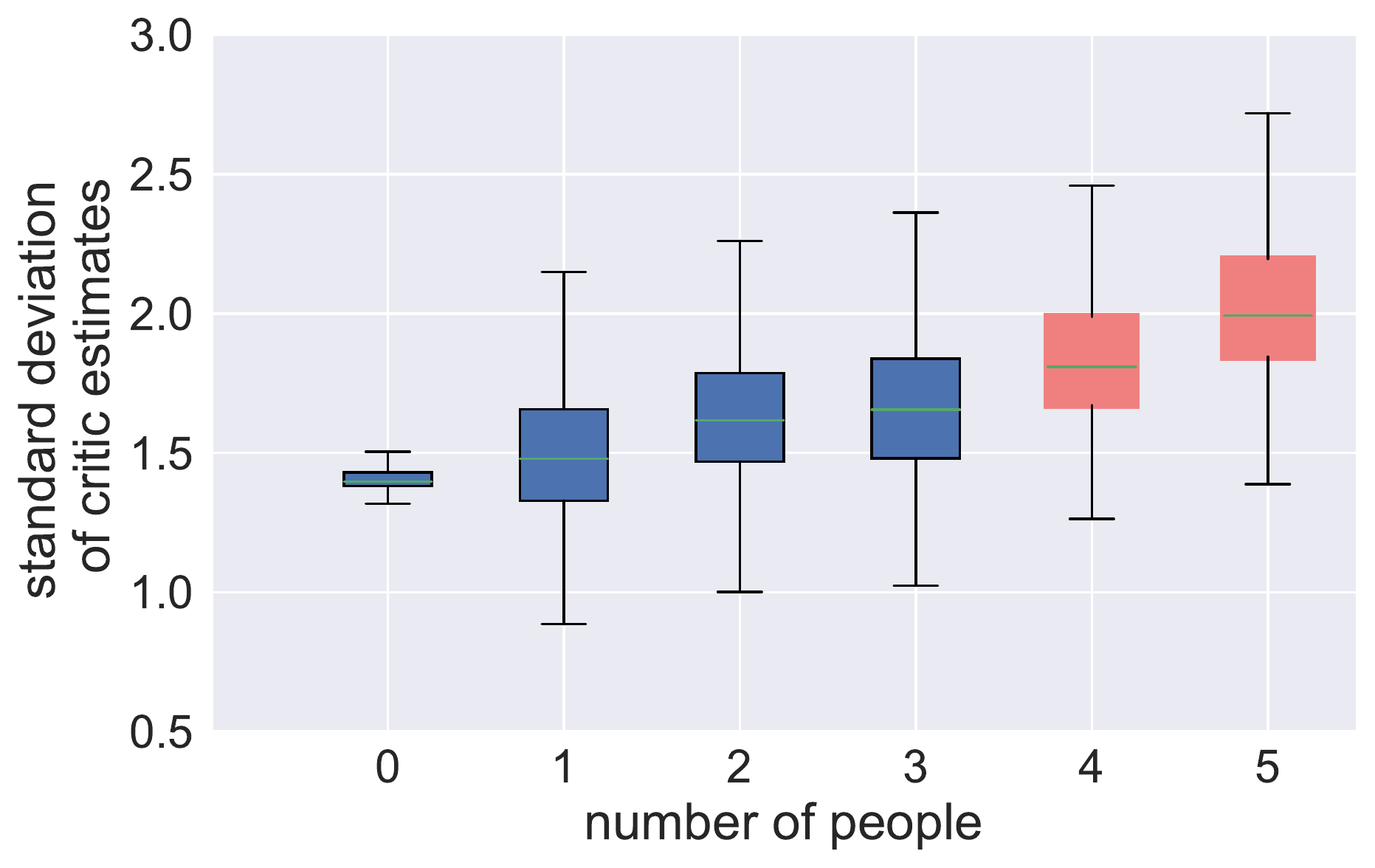}
         \caption{Standard deviation Distribution of critic estimates on test environments.}
         \label{fig:critic_std}
     \end{subfigure}
 \caption{Distributions of critic estimates on test environments. The OOD scenarios are highlighted in \textcolor{red}{red}.}
\label{fig:critic_dists}
\end{figure} 

\newpage
\section{Conclusion} \label{sec:results}
This paper presents an approach that utilizes context priors for an uncertainty-based \ac{mrl}. This is used for domain generalization and \ac{ood} detection. In order to assess the performance of our contribution, we benchmark it on a radar-tracking dataset towards related \ac{mrl} algorithms. In a set of multi-target radar-tracking scenarios, the proposed method outperforms related \ac{mrl} approaches in peak performance by $16\%$ and the baseline by $35\%$ while detecting \ac{ood} data with an F1-Score of 72\%. This shows that our method is more robust to environmental changes and well-addresses the \ac{ood} tracking scenarios. In future work, we want to expand the \ac{mrl} framework to different radar positions, radar devices, and internal settings (e.g., bandwidth, sampling frequency). \newpage



\bibliographystyle{IEEEtran}
\bibliography{references}

\end{document}